\documentclass{article} 
\usepackage[preprint]{colm2026_conference}

\usepackage[T1]{fontenc}
\usepackage[utf8]{inputenc}

\usepackage{microtype}
\usepackage{hyperref}
\usepackage{url}
\usepackage{booktabs}
\usepackage{graphicx}
\usepackage{subcaption}
\usepackage{algorithm}
\usepackage{algpseudocode}
\usepackage{listings}
\usepackage{amsmath}

\usepackage{booktabs}
\usepackage{pifont}

\newcommand{\cmark}{\ding{51}} 
\newcommand{\xmark}{\ding{55}} 

\usepackage{xcolor} 

\usepackage{lineno}

\definecolor{darkblue}{rgb}{0, 0, 0.5}
\hypersetup{colorlinks=true, citecolor=darkblue, linkcolor=darkblue, urlcolor=darkblue}

\title{LEAD: Breaking the No-Recovery Bottleneck in Long-Horizon Reasoning}

\author{Denys Pushkin \& Emmanuel Abbé \\
EPFL, Apple\\
\texttt{\{dpushkin, e\_abbe\}@apple.com} \\
}

\begin{document}

\ifcolmsubmission
\linenumbers
\fi

\maketitle

\begin{abstract}
Long-horizon execution in Large Language Models (LLMs) remains unstable even when high-level strategies are provided. Evaluating on controlled algorithmic puzzles, we demonstrate that while decomposition is essential for stability, extreme decomposition creates a ``no-recovery bottleneck''. We show that this bottleneck becomes critical due to highly non-uniform error distribution, where consistent errors on a few "hard" steps become irreversible.

To address this, we propose Lookahead-Enhanced Atomic Decomposition (LEAD). By incorporating short-horizon future validation and aggregating overlapping rollouts, LEAD provides enough isolation to maintain stability while retaining enough local context to correct errors. This enables the \texttt{o4-mini} model to solve Checkers Jumping up to complexity $n=13$, whereas extreme decomposition fails beyond $n=11$.
\end{abstract}

\section{Introduction}

Large language models (LLMs) have demonstrated impressive performance on many reasoning benchmarks, yet their accuracy degrades rapidly on tasks that require executing long sequences of reasoning steps, even when each individual step is simple.
This failure mode is especially pronounced in long-horizon algorithmic and puzzle-based tasks, where difficulty scales primarily with execution length rather than per-step complexity.

Prior works documented this phenomenon across diverse settings, including compositional factual queries \citep{press2023measuring}, mathematical reasoning \citep{hosseini2024not, zhou2025gsm}, and structured puzzle tasks \citep{shojaee2025illusion}.
A consistent finding in recent literature is the significant discrepancy between the success probability of a composed task and the product of marginal probabilities of isolated subtasks—a phenomenon termed the \emph{compositionality gap} \citep{press2023measuring}.
Notably, this gap persists across different model scales \citep{press2023measuring}, suggesting that while scaling improves individual step accuracy, it does not inherently resolve the compounding error rates of long-horizon execution.

Recent works indicate that the failures in long-horizon reasoning often stem from execution errors rather than inability to plan. For example, \citet{shojaee2025illusion} show that providing an explicit solution strategy yields little improvement on algorithmic puzzle tasks, while \citet{opus2025illusion} report that the very same models can generate a Python function that produces the complete solution to the puzzle, suggesting that high-level planning and algorithmic understanding remain intact.
Taken together, these findings motivate a clear separation between \emph{planning} and \emph{execution}, and call for a focused investigation of execution reliability in isolation.

Reliable execution over long horizons is critical in many real-world applications, including program synthesis and refactoring, tool-using agents, and proof translation into formal language.
In these settings, the high-level plan is often trivial or explicitly provided, yet failures arise from compounding errors in executing long sequences of simple, interdependent operations.

In this work, we aim to systematically understand the key principles behind robust long horizon execution. We follow the phylosophy of \cite{shojaee2025illusion} and study LLM's native abilities, prohibiting them to use any external tools.
We start by distilling common techniques into two fundamental motifs: context truncation (periodically restarting the generation with only the essential summary required to proceed) and task decomposition. Specifically, we focus on Atomic Decomposition - an extreme
form of decomposition where each step is executed in a separate model call using a minimal-context prompt.

We evaluate these motifs on Checkers Jumping and Tower of Hanoi, two algorithmic puzzles with adjustable complexity and known optimal strategy, previously proposed as benchmarks by \cite{shojaee2025illusion}.
By providing the solution strategy in the prompt, we explicitly isolate execution from planning.
These domains allow us to smoothly scale task complexity and analyze execution failures in a controlled setting, while serving as proxies for broader long-horizon reasoning challenges.

Our main contributions are summarized as follows:

\begin{itemize}

\item \textbf{The Necessity of Decomposition:} We demonstrate that for long-horizon execution, structural task decomposition is a prerequisite for stability. By comparing \emph{Context Truncation} with \emph{Atomic Decomposition}, we show that the deliberate isolation of reasoning steps—rather than mere context length management—is the primary driver of reliable execution across diverse models and benchmarks.

\item \textbf{No-Recovery Bottleneck of Strict Decomposition:} While decomposition is necessary, we identify a fundamental limitation of extreme decomposition—where every step is executed in complete isolation: its memoryless design makes local errors irreversible. We show that this structural limitation is critical due to a highly non-uniform error distribution, where errors are concentrated on a few "hard" steps. Once a model becomes consistently wrong on even a single step, success becomes statistically impossible—a bottleneck that persists even with majority voting—despite the model remaining highly competent across the rest of the horizon.

\item \textbf{Lookahead-Enhanced Atomic Decomposition (LEAD):} We propose LEAD, a framework that identifies the "Goldilocks zone" of task decomposition. By incorporating short-horizon future validation and aggregating overlapping rollout predictions, LEAD provides enough isolation to maintain stability while retaining enough local context to correct errors. This enables the \texttt{o4-mini} model to solve Checkers Jumping up to complexity $n=13$, whereas strict decomposition fails beyond $n=11$.

\end{itemize}

\section{Related work}

\paragraph{Long-Horizon Reasoning Failures in LLMs.}
A growing body of work documents systematic performance degradation of LLMs on long-horizon tasks. One line of research attributes failures to degradation of model's reasoning under long context conditioning ~\citep{liu2024lost, veseli2025positional, du2025context}. 
Complementary work shows degradation under compositionality or multi-step reasoning setting. \citet{press2023measuring} introduce the \emph{compositionality gap}, defined as the discrepancy between the success probability of a composed task and the product of success probabilities of its subtasks in isolation. Such gaps have been observed in two-hop factual composition~\citep{press2023measuring}, mathematical reasoning~\citep{hosseini2024not, zhou2025gsm}, and algorithmic puzzle tasks~\citep{shojaee2025illusion}. Notably, the compositionality gap does not diminish with model scaling~\citep{press2023measuring}, suggesting that reliable long-horizon execution does not emerge naturally through scale alone.

\paragraph{Context Management and Decomposition in LLM Reasoning.}
Prior work addresses long-horizon reasoning through two complementary strategies: managing context growth and decomposing tasks into smaller components.

One line of research mitigates long-context degradation by explicitly reducing effective prompt length. Retrieve-then-reason frameworks first select relevant information and then perform reasoning over a shortened context~\citep{li2024alr, du2025context}. More generally, several agent architectures adopt bounded-context interaction loops, carrying forward only a compressed or structured representation of prior state to prevent unbounded context growth~\citep{aghajohari2025markovian, liu2025context}. 

A complementary line of work focuses on task decomposition. Prompting strategies such as Least-to-Most~\citep{zhou2022least}, Plan-and-Solve~\citep{wang2023plan}, and Self-Ask~\citep{press2023measuring} decompose complex tasks into intermediate subproblems, typically executed within a single response that retains the full reasoning trace. Tree of Thoughts~\citep{yao2023tree} extends this paradigm through tree-structured exploration using multiple model calls, emphasizing planning and search. Multi-agent systems further distribute subtasks across specialized agents and aggregate their outputs~\citep{zhang2025agentorchestra, wang2025megaagent, meyerson2025solving}.

\paragraph{Solving Algorithmic Puzzles.}
Algorithmic puzzle domains have recently emerged as controlled testbeds for studying long-horizon reasoning and execution. \citet{shojaee2025illusion} introduce such puzzle benchmarks and demonstrate that model performance degrades sharply as problem complexity increases. Moreover, explicitly specifying solution plan in the prompt does not give any significant improvement. 
~\citet{sinha2509illusion} argue that this degradation can be attributed to self-conditioning on earlier mistakes. 
Recent work by \citet{meyerson2025solving} makes a major breakthrough in solving Tower of Hanoi puzzle using an extreme form of task decomposition with micro-agents. Their framework is very similar to the \textit{Atomic Decomposition}, which we will consider in Section \ref{sec:baselines}, but augmented with a ``first-to-ahead-by-k'' voting scheme and a red-flagging mechanism for suspicious sample filtering. While their work demonstrates an impressive results on Tower of Hanoi, we will show that their approach suffers from ``no-recovery bottleneck'' and yields only limited gains on Checkers Jumping Puzzle, where the steps are natively harder. 

\section{Tasks}
We use two algorithmic puzzles as our testbed: \emph{Tower of Hanoi} and \emph{Checkers Jumping}. \textbf{Tower of Hanoi} requires moving $n$ disks between three pegs such that no larger disk is placed on a smaller one, and the optimal solution takes $2^n - 1$ moves. \textbf{Checkers Jumping} involves swapping two blocks of $n$ colored pieces on a 1D board via slides and jumps, and can be solved in $(n+1)^2 - 1$ moves. You can find a formal task specifications and solution strategies in the prompt examples provided in Appendix \ref{sec:prompts}.

For both tasks, we provide the optimal strategy in the prompt and require the model to generate the full sequence of moves and resulting states without external tools. This structured format (see Listings~\ref{lst:task1}, \ref{lst:task2}) allows us to distinguish between errors in move selection and state execution, enabling a fine-grained error analysis.

\begin{figure}[tb]
    \centering

    \begin{subfigure}{\textwidth}
        \centering
        \includegraphics[width=\textwidth]{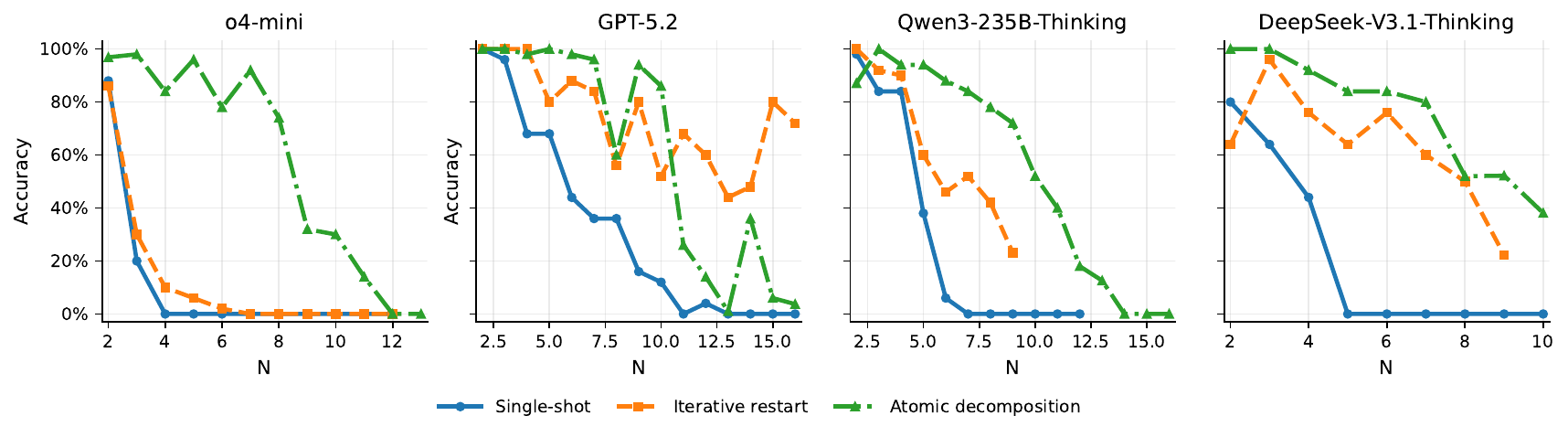}
        \caption{Checkers Jumping}
        \label{fig:checkers-methods-comparison}
    \end{subfigure}

    \vspace{1em} 

    \begin{subfigure}{\textwidth}
        \centering
        \includegraphics[width=0.67\textwidth]{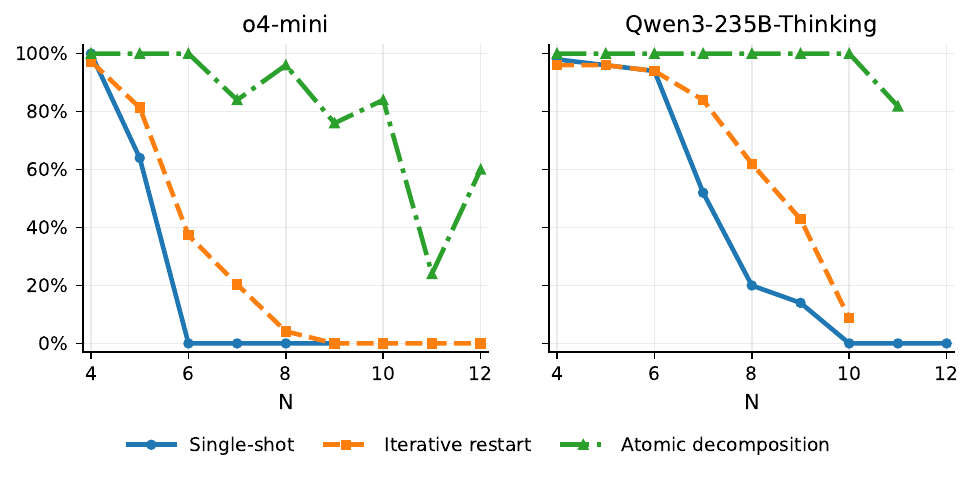}
        \caption{Tower of Hanoi}
        \label{fig:tower-methods-comparison}
    \end{subfigure}

    \caption{Comparison of baseline methods across two puzzles.}
    \label{fig:methods-comparison}
\end{figure}

\section{Baseline methods} \label{sec:baselines}

We study execution strategies for long-horizon reasoning under a fixed prompt prefix that specifies the puzzle description and its optimal solution strategy. Across all methods, only the task instance (e.g., problem size $n$ and current state) varies.


\paragraph{Single-shot generation.}
In the single-shot setting, the model is prompted to generate the entire sequence of steps---including both moves and resulting states---in a single response. All intermediate reasoning and execution history remain within the same context window.

\paragraph{Iterative restart.}

In the iterative restart setting, the model is repeatedly prompted with the current state and asked to complete as many steps as it can. Within each response, the model may generate multiple consecutive steps until it either completes the task or reaches its capacity (as decided by the model itself). In the latter case, the model outputs the partial solution, and the updated state is fed back to the model in a new prompt.
Iterative restart prevents unbounded context growth, but it's generally does not lead to the minimal-context execution since the model can solve multiple steps within a single response, thus accumulating the history of previous steps solution in reasoning traces.

\paragraph{Atomic Decomposition.} Each reasoning step is executed in isolation: at step $i$, the model receives only the current state $s_i$, and prompted to generate one move and next state $s_{i+1}$. By discarding prior history and intermediate traces, this approach eliminates context-based error accumulation and treats each step as an independent atomic operation.

\section{Lookahead-Enhanced Atomic Decomposition (LEAD)} \label{sec:lead}

Despite mitigating context degradation, Atomic Decomposition lacks a recovery mechanism: since the steps are independent and history is discarded, error at any step becomes irreversible. It creates a \textit{no-recovery bottleneck}, and motivates for augmenting atomic execution with some error-recovery mechanism without reintroducing large context dependencies.

To address the no-recovery bottleneck, we propose \emph{Lookahead-Enhanced Atomic Decomposition (LEAD)}. This framework stabilizes execution by combining short-horizon future validation with consistency-filtering and aggregation across overlapping rollouts.

\textbf{Lookahead mechanism.} At step $i$, instead of predicting only the immediate next state, the model generates a rollout of $k$ future steps: $(s_i \rightarrow s_{i+1} \rightarrow \dots \rightarrow s_{i+k})$. By forecasting future states, the model can implicitly detect and revise inconsistencies; if a locally incorrect decision leads to future contradictions, the model may self-correct within the same rollout. This introduces forward validation while maintaining a bounded context length.

\textbf{Overlapping aggregation.} LEAD aggregates predictions across rollouts initiated from the current state $i$ and the previous $h-1$ states. For each rollout, we enforce consistency with the committed trajectory by resampling until all predicted steps overlapping with the history match the already executed steps. This \textit{positive conditioning} (Section \ref{sec:pos_cond}) ensures the model operates under the assumption of a correct prior path, significantly improving rollout accuracy. 

\textbf{Voting procedure.} To optimize for computational efficiency, LEAD employs a two-tier verification process (Algorithm~\ref{alg:lead-compact}). For each step, the model first generates $v$ independent atomic predictions for the current state $s_i$. If all samples reach a consensus, the model proceeds via a \textit{fast-path}, accepting the step without further computation. 

However, if disagreement is detected among the base samples—signaling a potential "hard step"—the framework escalates to the full lookahead mechanism. Predictions for the current state are then extracted from overlapping rollouts and aggregated via a margin-based voting procedure \citep{meyerson2025solving}: voting continues until one candidate outvotes all others by a margin $t$. This hierarchical structure ensures that the heavy computational cost of generating $k$-step rollouts is only incurred when necessary to resolve local instabilities. This process effectively smooths the error distribution: difficult steps that cause atomic disagreement are often correctly resolved when viewed through the wider context of rollouts originating from nearby states.

The descriptive version of the LEAD algorithm is presented in Algorithm \ref{alg:lead-compact}, and you can find the formal version of the algorithm in Appendix \ref{app:algo}.

\begin{algorithm}[h]
\caption{LEAD (Lookahead-Enhanced Atomic Decomposition)}
\label{alg:lead-compact}
\begin{algorithmic}[1]
\Require Model $M$, votes $v$, depth $k$, history $h$, margin $t$, initial state $x$
\Ensure Action sequence $A$

\State $A \gets [\,]$

\While{not finished}
    \State \Comment{Sample $v$ independent next-step predictions for current state $x$}
    \State $\{(a^{(m)}, x^{(m)})\}_{m=1}^v \sim M(\text{current state } x)$
    
    \If{all samples agree}
        \State $(a, x) \gets (a^{(1)}, x^{(1)})$ \Comment{Fast-path: accept consensus}
    \Else
        \State $V[\cdot] \gets 0$ \Comment{Initialize vote counter}
        \While{$\max V - \text{2nd-max } V < t$}
            \For{$x'$ in last $h$ states}
                \State Sample a $k$-step rollout from $x'$ (consistent with history)
                \State Let $(a, x)$ be the step in this rollout corresponding to the current time
                \State $V[(a, x)] \mathrel{+}= 1$
            \EndFor
        \EndWhile
        \State $(a, x) \gets \arg\max V$ \Comment{Resolve disagreement via lookahead consensus}
    \EndIf
    
    \State $A \gets A \cup \{a\}$
\EndWhile

\State \Return $A$
\end{algorithmic}
\end{algorithm}



    
    


\section{Experiments}

\paragraph{Setup.} We evaluate o4-mini, GPT-5.2 (both at low reasoning effort), Qwen3-235B-Thinking, and DeepSeek-V3.1-Thinking models. Results are averaged over 50 independent runs, with the exception of DeepSeek-V3.1-Thinking, and Tower of Hanoi baselines (Atomic Decomposition and Iterative Restart), which use 25 runs due to computational cost. Detailed prompts are provided in Appendix \ref{sec:prompts}.

\subsection{Decomposition is necessary for long-horizon stability.} Figure~\ref{fig:methods-comparison} shows that Atomic Decomposition generally outperforms other baselines, confirming that isolating reasoning steps is essential for stability even under strict context truncation. A notable exception is GPT-5.2, where Iterative Restart performs better at higher complexities. We attribute this to the model's strong \textit{positive-conditioning} effect (Section~\ref{sec:pos_cond}): by restarting at varying positions, the model can initiate generation from "easy" states. As shown in Figure~\ref{fig:lookahead-positions-d}, this allows GPT-5.2 to solve subsequent hard steps that it otherwise fails to resolve when they are positioned at the start of a generation.

\subsection{The hard-step bottleneck} \label{sec:hard-step-bottle}

To understand why decomposition fails as complexity increases, we analyze per-step error distributions (see Appendices ~\ref{app-a}, \ref{app-b} for qualitative analysis). We find a fundamental divergence in difficulty between the two tasks. In \textbf{Tower of Hanoi}, error probabilities remain uniformly low ($\leq 0.08$ for \texttt{o4-mini} up to $n=12$), explaining why stepwise execution with majority voting scale effectively \citep{meyerson2025solving}. 

\begin{figure}[t]
    \centering
    \begin{subfigure}[b]{0.58\columnwidth}
        \centering
        \includegraphics[width=\textwidth]{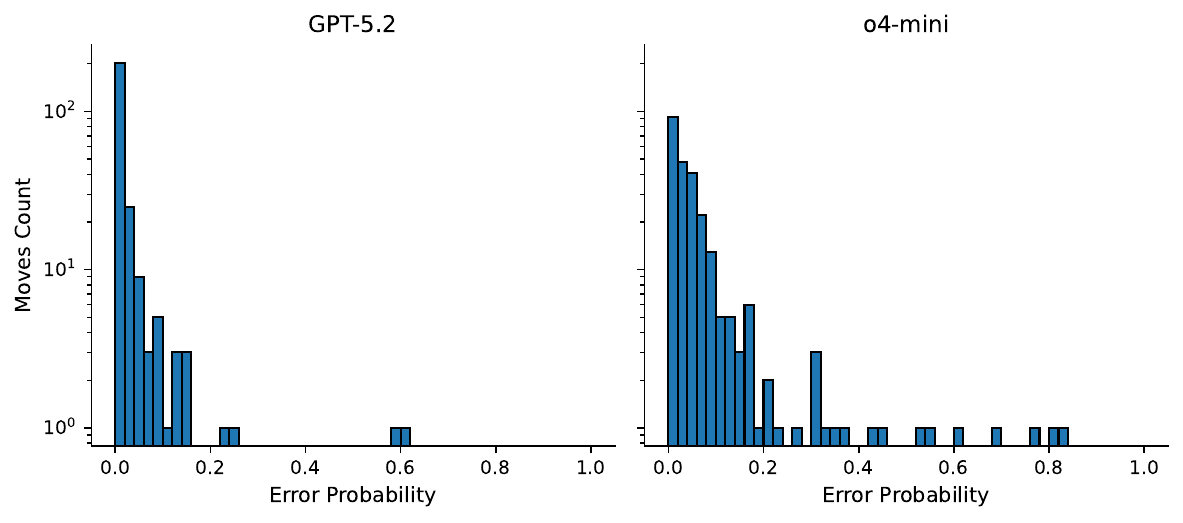}
        \caption{Within-model: step error histogram.}
        \label{fig:error-prob-hist}
    \end{subfigure}
    \hfill
    \begin{subfigure}[b]{0.39\columnwidth}
        \centering
        \includegraphics[width=\textwidth]{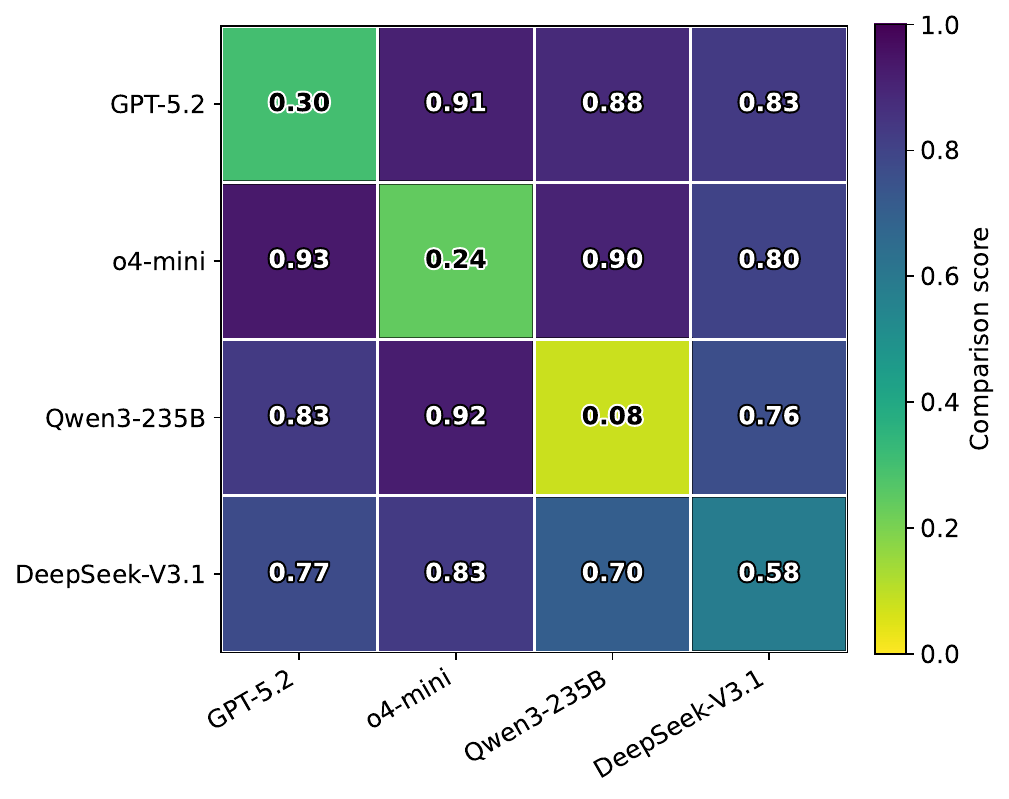}
        \caption{Across-model: TV distance.}
        \label{fig:error-dist-heatmap}
    \end{subfigure}
    
    \caption{\textbf{Non-uniform error distributions in Checkers Jumping.} 
    (a) Per-step error histograms for o4-mini and GPT-5.2 ($n=15$) show that failures are concentrated on specific ``hard steps.'' 
    (b) Pairwise Total Variation (TV) distance heatmap ($n=13$) between error distributions of different models. High off-diagonal values indicate that hard steps vary significantly across architectures, while low diagonal values confirm the robustness of our estimates.}
    \label{fig:combined-error-analysis}
\end{figure}

Conversely, \textbf{Checkers Jumping} exhibits a highly non-uniform distribution. While most steps have near-zero error, a subset of ``hard steps'' emerges where error probabilities exceed 0.5 for \texttt{o4-mini} ($n \geq 12$) and \texttt{GPT-5.2} ($n = 15$), as shown in Fig.~\ref{fig:error-prob-hist}. This concentration creates a \textit{no-recovery bottleneck}: because errors are consistent rather than independent, naïve majority voting fails regardless of the number of votes, stalling the model at these critical junctions.

\paragraph{Model-specific heterogeneity.} Our analysis suggests these hard steps are often model-specific rather than intrinsic to the task. We quantify this by calculating the pairwise Total Variation (TV) distance between empirical error distributions of different models on Checkers Jumping ($n=13$). As shown in Fig.~\ref{fig:error-dist-heatmap}, distributional divergence between different models significantly exceeds the internal variance of any single model. This indicates that different architectures struggle with \textbf{largely distinct} subsets of the state space. Such heterogeneity suggests that the "hard-step" barrier is not a universal limit of the task, but a behavioral signature of specific models, implying that ensembling or adaptive selection could further stabilize long-horizon execution.

\begin{figure}[tb]
  \centering

  \begin{subfigure}{0.48\columnwidth}
    \centering
    \includegraphics[width=\linewidth]{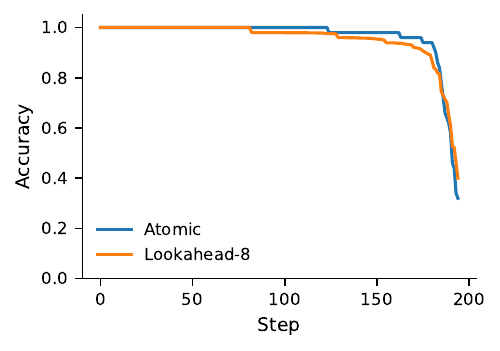}
    \caption{o4-mini}
  \end{subfigure}
  \hfill
  \begin{subfigure}{0.48\columnwidth}
    \centering
    \includegraphics[width=\linewidth]{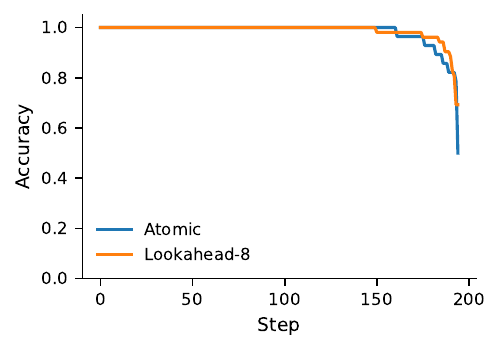}
    \caption{GPT-5.2}
  \end{subfigure}

  \caption{\textbf{Lookahead improves robustness on hard steps.} Rank-ordered step accuracies for Atomic decomposition vs. Lookahead ($k=8$) show that while Lookahead introduces a slight overhead on ``easy'' steps for the o4-mini model, it consistently boosts performance on the hardest moves (the right tail) for both tested models. The step accuracies were estimated by sampling 50 solutions for each step in Checkers Jumping ($n=13$).}

  \label{fig:lookahead-dist}
\end{figure}

\subsection{Mechanisms of lookahead recovery} \label{sec:pos_cond}

Lookahead (Section~\ref{sec:lead}) functions as a targeted error-correction tool. As shown in Figure~\ref{fig:lookahead-dist}, while a $k=8$ rollout may slightly degrade average-case accuracy for \texttt{o4-mini}, it consistently improves performance on the ``hardest'' steps for both \texttt{o4-mini} and \texttt{GPT-5.2}. This confirms our intuition: while forecasting adds complexity to trivial steps, it provides a critical corrective signal at high-entropy junctions where atomic intuition fails (see App.~\ref{app-c}).

\begin{figure*}[b]
  \centering



  \begin{subfigure}{0.48\columnwidth}
    \centering
    \includegraphics[width=\linewidth]{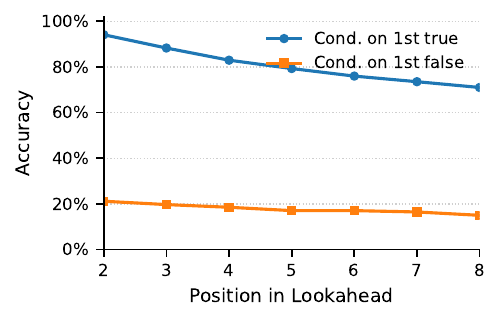}
    \caption{o4-mini} \label{fig:lookahead-positions-c}
  \end{subfigure}
  \hfill
  \begin{subfigure}{0.48\columnwidth}
    \centering
    \includegraphics[width=\linewidth]{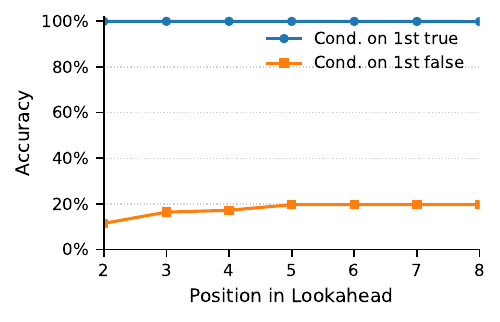}
    \caption{GPT-5.2} \label{fig:lookahead-positions-d}
  \end{subfigure}

  \caption{Conditional accuracy of length-8 rollouts for Checkers Jumping ($n=13$). The plot demonstrates how lookahead rollout accuracy evolves when conditioned on the correctness of the first predicted step. Data is aggregated from 50 samples per step across the entire horizon (excluding the last 7 positions).}
  \label{fig:lookahead-positions}
\end{figure*}

The efficacy of this signal is governed by two competing phenomena. First, \textbf{positional accuracy decay}: for \texttt{o4-mini}, accuracy typically degrades as the prediction moves further from the current state (Figure~\ref{fig:lookahead-positions-c}). Second, a powerful \textbf{self-conditioning effect} can counteract this decay. For \texttt{GPT-5.2}, if the initial rollout step is correct, the model ``locks onto'' a valid trajectory, maintaining high stability across the horizon (Figure~\ref{fig:lookahead-positions-d}). This effect is also present in \texttt{o4-mini}, albeit more weakly. LEAD exploits this by enforcing consistency with committed steps, ensuring rollouts benefit from this positive conditioning.

\subsection{Main results: surpassing the hard-step-bottleneck}

Table~\ref{tab:lead_results} shows that LEAD consistently outperforms both vanilla Atomic Decomposition and its majority-voting variant across the critical complexity thresholds where standard methods collapse. By neutralizing the ``hard-step'' failures that cripple decomposition, LEAD extends the reliable execution horizon and achieves a new state-of-the-art on the Checkers Jumping benchmark. These results demonstrate that while minimal context is often prioritized for stability, a hybrid approach—selectively incorporating the self-corrective signal of lookahead rollouts (Section~\ref{sec:lead})—is essential to navigate irreversible failure points in long-horizon reasoning.


Crucially, the failure of Atomic Decomposition with Voting (AD + Voting) reveals a fundamental limitation: while majority voting over atomic steps is effective for tasks like Tower of Hanoi, where error probabilities remain uniformly low, it is insufficient for tasks like Checkers Jumping which exhibit highly non-uniform error distributions. On these ``hard steps,'' errors are not merely stochastic but systematic; consequently, simply increasing the voting budget for AD + Voting fails to improve performance, as the model consistently converges on the same incorrect consensus. This also highlights a potential limitation of the MAKER framework \citep{meyerson2025solving}. Although their framework builds upon AD + Voting by incorporating the ``red-flagging'' of suspicious responses, it remains unclear to what extent such a mechanism can effectively identify or correct systematic errors where the model's internal confidence in an incorrect state remains high.

For these experiments, we configured LEAD with parameters $h=2$, $k=8$, and $t=3$, setting the base votes $v$ equal to the task complexity $n$. Ablations using a larger history window ($h=3$) yielded no significant performance gains (see Table \ref{tab:ablation}). To ensure a rigorous comparison, the Atomic Decomposition baseline was equipped with a competitive ``first-to-ahead-by-3'' voting scheme \citep{meyerson2025solving}. Despite this, LEAD's cross-step aggregation provided a superior error-correction signal, effectively ``rescuing'' trajectories that otherwise hit the $n=12$ bottleneck when using \texttt{o4-mini}.

A notable exception occurred during the evaluation of GPT-5.2 at $n=17$ and Qwen3-235B-Thinking at $n=16$, where maintaining the $v=n$ strategy would result in near-zero accuracy due to frequent false consensus during the base voting procedure (as visualised by error probability histograms in Figure \ref{fig:error-hist-gpt52} and Figure \ref{fig:error-hist-qwen}). To address this while keeping the experiment computationally tractable, we modified the sampling strategy by pre-generating a pool of 70 independent predictions for each step. During LEAD evaluation, we then sampled $v=70$ candidates with replacement from this fixed pool for the base voting stage. This configuration significantly reduced the probability of an incorrect "fast-path" agreement, effectively shifting the primary decision-making responsibility to the more robust lookahead-based voting phase.






\begin{table}[t]
  \centering
  \small

  \begin{tabular}{lccc c ccccc c ccc}
    \toprule
    & \multicolumn{3}{c}{\textit{o4-mini} ($N$)} & 
    & \multicolumn{5}{c}{\textit{GPT-5.2} ($N$)} & 
    & \multicolumn{3}{c}{\textit{Qwen} ($N$)} \\
    \cmidrule(lr){2-4} \cmidrule(lr){6-10} \cmidrule(lr){12-14}
    Method  
    & 11 & 12 & 13 
    & 
    & 13 & 14 & 15 & 16 & 17 
    & 
    & 14 & 15 & 16 \\
    \midrule
    AD   
         & 14 & 0  & 0  
         & & 0   & 36  & 0   & 4   & 0  
         & & 0   & 0   & 0 \\
    AD + Voting  
         & 76 & 16 & 0  
         & & 20  & \textbf{100} & 0 & 96  & 0  
         & & \textbf{100} & 94  & 0 \\
    LEAD  
         & \textbf{100} & \textbf{56} & \textbf{76} 
         & & \textbf{100} & \textbf{100} & \textbf{100} & \textbf{100} & \textbf{80} 
         & & 84  & \textbf{96}  & \textbf{60} \\
    \bottomrule
  \end{tabular}

  \caption{\textbf{LEAD scales reasoning performance across architectures.} Success rates (\%) on Checkers Jumping comparing Atomic Decomposition (AD), AD with first-to-ahead-by-3 voting, and LEAD (with parameters $k=8, h=2, t=3$). Bold indicates best performance per horizon.}
  \label{tab:lead_results}
\end{table}

\subsection{Inference cost analysis}

\begin{figure}[t]
  \centering

  \begin{subfigure}{0.48\columnwidth}
    \centering
    \includegraphics[width=\linewidth]{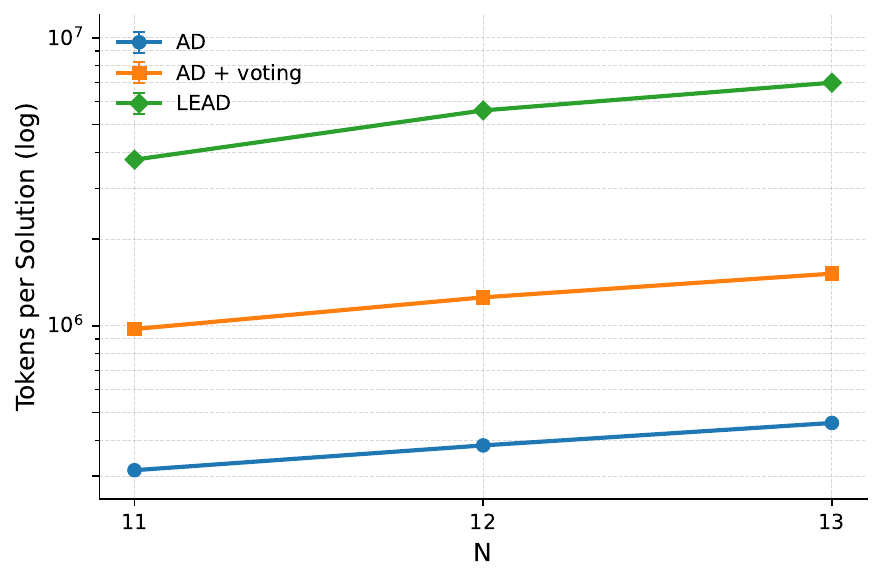}
    \caption{o4-mini}
  \end{subfigure}
  \hfill
  \begin{subfigure}{0.48\columnwidth}
    \centering
    \includegraphics[width=\linewidth]{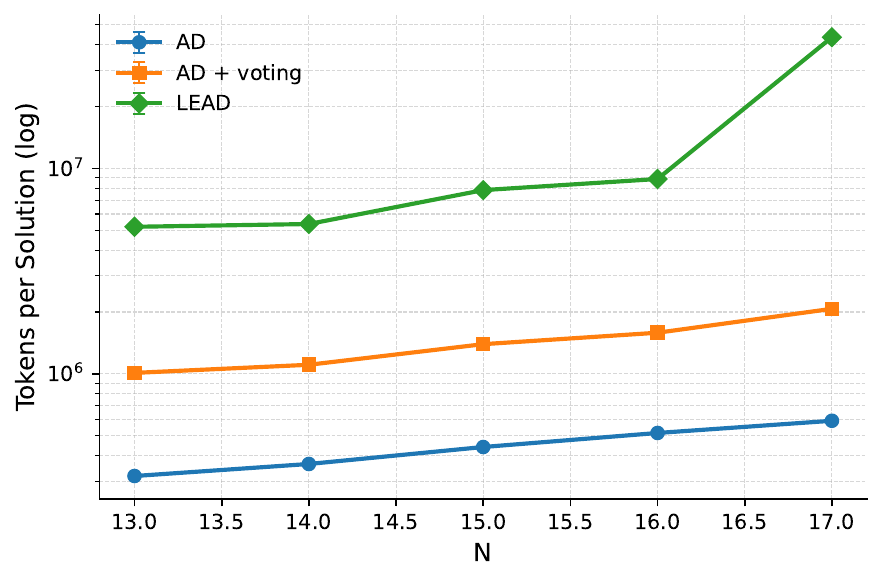}
    \caption{GPT-5.2}
  \end{subfigure}

  \caption{\textbf{Computational cost of robust execution.} Average number of tokens per solution for Atomic Decomposition (AD), AD with first-to-ahead-by-3 voting, and LEAD (parameters as in Table~\ref{tab:lead_results}). While LEAD increases token consumption, it enables the model to surpass the no-recovery bottleneck that typically limits extreme decomposition techniques. Notably, simply scaling the budget for AD + Voting is unlikely to yield similar gains, as errors on ``hard steps'' are often systematic rather than stochastic, rendering simple majority voting ineffective.}
  \label{fig:tokens}
\end{figure}

Figure~\ref{fig:tokens} illustrates the average token count per solution for the methods in Table~\ref{tab:lead_results}. While LEAD is computationally more expensive than the baselines, it provides a unique capability: extending the reliable execution horizon beyond the "hard-step" threshold. 

As established in Section~\ref{sec:hard-step-bottle}, scaling via simple stepwise voting fails once a model's error distribution becomes consistent at critical junctions. Any decomposition that executes steps in total isolation eventually hits this "no-recovery" wall. To our knowledge, LEAD is the first algorithm to surpass this threshold by replacing isolated steps with overlapping rollouts. While we prioritized performance over inference efficiency, the per-solution token estimates (Appendix~\ref{app:complexity}) provide a baseline for future research into more efficient paradigms for scaling test-time compute in long-horizon tasks.

\subsection{Ablation Studies}

To evaluate the individual contributions of LEAD's core components—\textit{lookahead}, \textit{consistency filtering}, and \textit{overlapping aggregation}—we compare the full framework against ablated variants across the critical complexity ranges where standard Atomic Decomposition typically fails. In the ``no-lookahead'' baseline, rollouts are truncated to the minimum length required to aggregate predictions for the current step, effectively isolating the impact of future-state forecasting.

As shown in Table~\ref{tab:ablation}, optimal performance is achieved only when all three mechanisms are integrated, validating our architectural design. Our results indicate that while the filtering component provides incremental improvements, overlapping aggregation is the primary driver of robust performance, as it enables the cross-step error correction necessary to navigate high-complexity trajectories.

\begin{table*}[!b]
\centering
\small
\begin{tabular}{ccc ccc c ccc}
\toprule
Lookahead & Filtering & Aggregation 
& \multicolumn{3}{c}{o4-mini ($N$)} 
& 
& \multicolumn{3}{c}{GPT-5.2 ($N$)} \\
\cmidrule(lr){4-6} \cmidrule(lr){8-10}
          &           &             
& 11 & 12 & 13 
& 
& 13 & 14 & 15 \\
\midrule
\multicolumn{10}{c}{$h=3$} \\
\cmark & \cmark & \cmark 
& 96  & \textbf{56} & \textbf{84}
& 
& 100 & 100 & \textbf{100} \\
\xmark & \cmark & \cmark 
& \textbf{100} & 36 & 24
& 
& 100 & 100 & \textbf{100} \\
\cmark & \xmark & \cmark 
& 96  & 52 & 80
& 
& 100 & 100 & 92 \\
\cmark & \cmark & \xmark 
&  96 &  4  & 0
&
&  72  & 92 & 60 \\
\addlinespace
\multicolumn{10}{c}{$h=2$} \\
\cmark & \cmark & \cmark 
& 100 & 56 & \textbf{76}
& 
& 100 & 100 & \textbf{100} \\
\xmark & \cmark & \cmark 
& 100 & 32 & 28
& 
& 100 & 100 & 32 \\
\cmark & \xmark & \cmark 
& 100 & \textbf{64} & 52
& 
& 100 & 100 & 80 \\
\cmark & \cmark & \xmark 
&  96 &  4  & 0
&
&  72  & 92 & 60 \\
\bottomrule
\end{tabular}
\caption{\textbf{Ablation study of LEAD components.} Success rates (\%) for o4-mini and GPT-5.2 on Checkers Jumping across varying horizons $N$. We evaluate the contribution of Lookahead rollouts, Consistency Filtering, and Overlapping Aggregation (with window size $h$). Results show that removing any component leads to performance degradation, and Aggregation is the most critical component for maintaining performance. Parameters $k=8, t=3$ are held constant, and we set $v=N$ to adapt to task complexity. Note that the ``no-aggregation'' results are identical across both $h$ blocks as that parameter becomes irrelevant.}
\label{tab:ablation}
\end{table*}

\section{Conclusion}
In this work, we identified the no-recovery bottleneck as a primary obstacle to Atomic Decomposition in long-horizon execution. Our empirical analysis reveals that success is not limited by average model competence, but by a highly non-uniform error distribution where a few ``hard'' steps act as irreversible points of failure. We demonstrated that while minimal-context atomic execution provides stability for uniform tasks like the Tower of Hanoi, it lacks the necessary corrective signal to navigate the high-entropy junctions found in more complex puzzles.

To address this, we introduced LEAD (Lookahead-Enhanced Atomic Decomposition). By integrating temporal rollouts with consistency filtering and smoothed voting mechanism, LEAD identifies and corrects local errors before they propagate into global failures. Our results show that LEAD significantly extends the reliable reasoning horizon of frontier models like \texttt{o4-mini}, \texttt{Qwen3-235B-Thinking}, and \texttt{GPT-5.2}, outperforming existing stepwise baselines. Ultimately, our findings suggest that the next frontier in robust AI execution lies not in further context reduction, but in adaptive motifs that can selectively leverage lookahead to stabilize critical transitions.

\bibliography{ref.bib}
\bibliographystyle{colm2026_conference}

\newpage

\appendix

\section{Full algorithm description} \label{app:algo}

To formally define the \textsc{LEAD} framework, we introduce two prompt construction functions, $\phi_{\mathrm{AD}}$ and $\phi_{\mathrm{LA}}$. The function $\phi_{\mathrm{AD}}(x)$ maps a single state $x$ to a prompt following the Atomic Decomposition (AD) paradigm, instructing the model to predict exactly one next step $(a, x')$. The function $\phi_{\mathrm{LA}}(x, k)$ maps a state $x$ and a lookahead depth $k$ to a prompt instructing the model to generate a sequence of $k$ consecutive steps starting from $x$.

\begin{algorithm}[h]
\caption{LEAD (Lookahead-Enhanced Atomic Decomposition)}
\label{alg:lead}
\begin{algorithmic}[1]
\Require Model $M$, vote count $v$, lookahead depth $k$, history window $h \le k$, margin $t$, initial state $x_0$, horizon $S$
\Ensure Action sequence $A$, state sequence $X$

\State $A \gets [\bot]$, \quad $X \gets [x_0]$ \Comment{$\bot$ denotes a null action used only to align indices}
\State $x \gets x_0$

\For{$i = 1$ to $S$}
    \State $\{y^{(m)}\}_{m=1}^v \sim M\big(\phi_{\mathrm{AD}}(x)\big)$ \Comment{$y^{(m)} = (a^{(m)}, x^{(m)})$}
    
    \If{$y^{(1)} = \cdots = y^{(v)}$}
        \State $(a, x) \gets y^{(1)}$ \Comment{If atomic samples agree, execute directly}
    \Else
        \State $(a, x) \gets \Call{LookaheadVote}{M, A, X, h, k, t}$ \Comment{Resolve disagreement using lookahead aggregation}
    \EndIf
    
    \State Append $a$ to $A$
    \State Append $x$ to $X$
\EndFor

\State \Return $A$, $X$
\end{algorithmic}
\end{algorithm}

\floatname{algorithm}{Procedure}
\begin{algorithm}[h]
\caption{\textsc{LookaheadVote}}
\label{alg:lookahead-voting}
\begin{algorithmic}[1]
\Require Model $M$, history $(A, X)$, history window $h$, lookahead depth $k$, margin $t$
\Ensure Selected pair $y^\star = (a^\star, x^\star)$ for current step

\State $V[y] \gets 0$ for all candidate pairs $y$
\State $i \gets |X| - 1$ \Comment{Current state index; next decision is for step $i+1$}

\While{True}
    \For{$\ell = 0$ to $\min(h-1, i)$}
        \State $y^{(\ell)} \gets \Call{ConsistentRollout}{M, A, X, i-\ell, k}$ \Comment{Candidate induced by rollout started at state $x_{i-\ell}$}
        \State $V[y^{(\ell)}] \gets V[y^{(\ell)}] + 1$
    \EndFor
    
    \State $y^\star \gets \arg\max_y V[y]$
    \If{$V[y^\star] \ge t + \max_{y' \ne y^\star} V[y']$}
        \State \Return $y^\star$ \Comment{Return once one candidate leads by margin $t$}
    \EndIf
\EndWhile
\end{algorithmic}
\end{algorithm}

\begin{algorithm}[h]
\caption{\textsc{ConsistentRollout}}
\label{alg:get-lookahead}
\begin{algorithmic}[1]
\Require Model $M$, history $(A, X)$, rollout start index $i_0$, lookahead depth $k$
\Ensure Candidate pair for the next undecided step

\State $r \gets (|X| - 1) - i_0$ \Comment{Number of already-committed steps after rollout start}
\State Define $a_j \gets A[j]$, $x_j \gets X[j]$ for all $j = 0,\ldots,|X|-1$

\While{True}
    \State $(\hat{y}_{i_0+1}, \ldots, \hat{y}_{i_0+k}) \sim M\big(\phi_{\mathrm{LA}}(x_{i_0}, k)\big)$ \Comment{Sample a $k$-step lookahead rollout}
    
    \If{$\hat{y}_{i_0+j} = (a_{i_0+j}, x_{i_0+j})$ for all $j = 1,\ldots,r$}
        \State \Return $\hat{y}_{i_0+r+1}$ \Comment{Return the first prediction beyond the committed prefix}
    \EndIf
\EndWhile
\end{algorithmic}
\end{algorithm}

\clearpage

\section{Details on error analysis.} \label{app-a}


We distinguish two sources of failure: \emph{move finding} (selecting the correct action) and \emph{move execution} (correctly updating the state based on that action).
We found that for Checkers Jumping the errors are dominated by move execution failures, whereas for Tower of Hanoi errors arise primarily from incorrect move selection (see Figure~\ref{fig:error-types} in Appendix~\ref{app-a}).

Manual inspection suggests that execution errors in Checkers Jumping primarily arise from incorrect updates to long blocks of same-color checkers (see Appendix~\ref{app-b} for representative examples). This behavior aligns with known limitations of Transformer models in copying long sequences of identical symbols~\cite{zhou2023algorithms}. In contrast, the Tower of Hanoi representation assigns a unique numerical identifier to each disk, which makes state updates substantially easier for the model. We hypothesize that introducing unique identifiers for checkers (e.g., $R_1,\ldots,R_N,B_1,\ldots,B_N$) or explicitly enumerating board positions (e.g., $[1{:}R,2{:}R,\ldots,2n{:}B]$) could mitigate these errors. We leave a systematic investigation of these alternatives to future work.

\begin{figure}[hb]
    \centering

    \includegraphics[width=\textwidth]{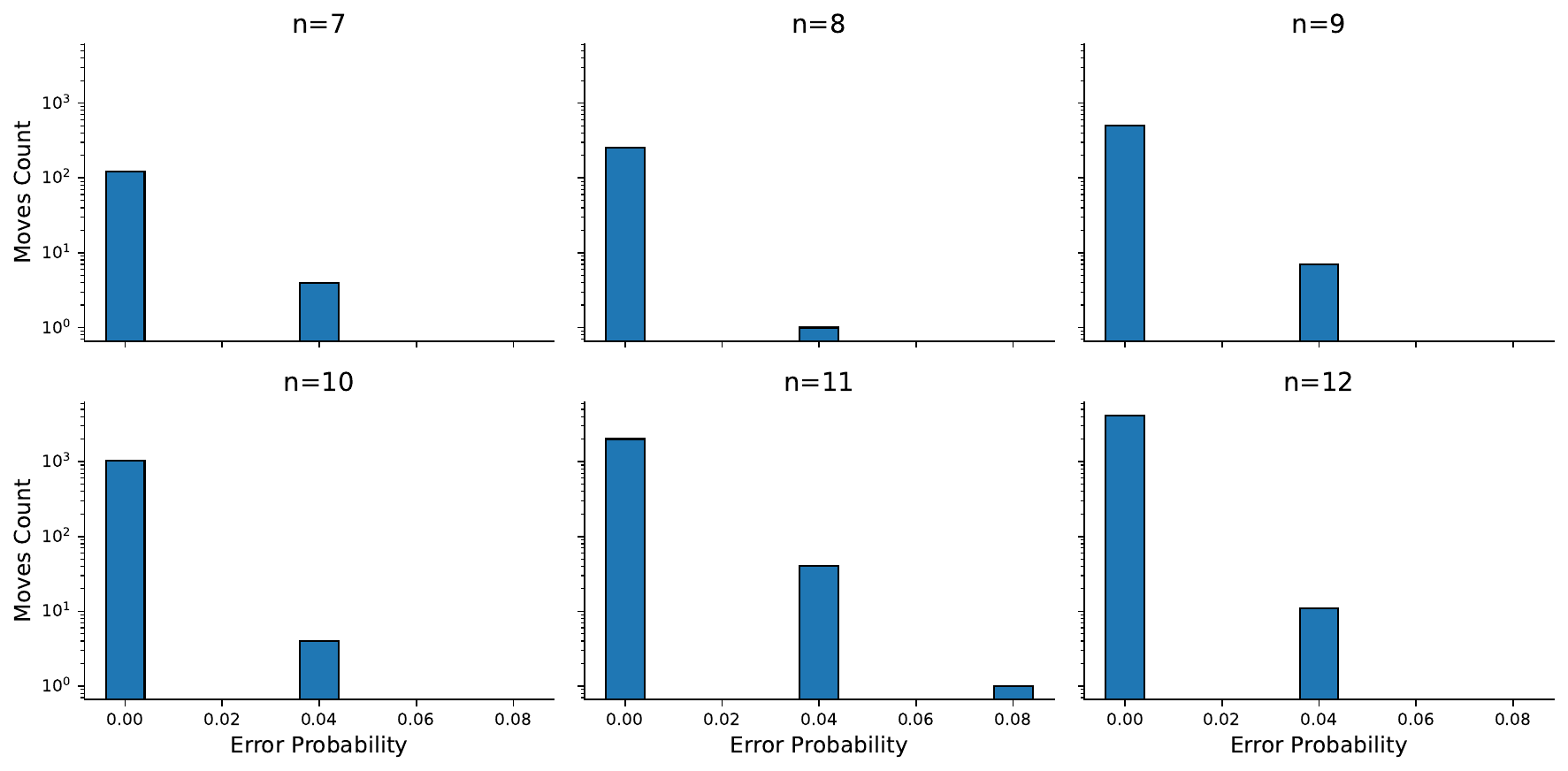}

    \caption{Error count histogram of o4-mini model for Tower of Hanoi. The error counts were estimated by sampling 25 independent solutions for each step.}
    \label{fig:error-hist}
\end{figure}

\clearpage

\begin{figure}[hb]
    \centering

    \includegraphics[width=\textwidth]{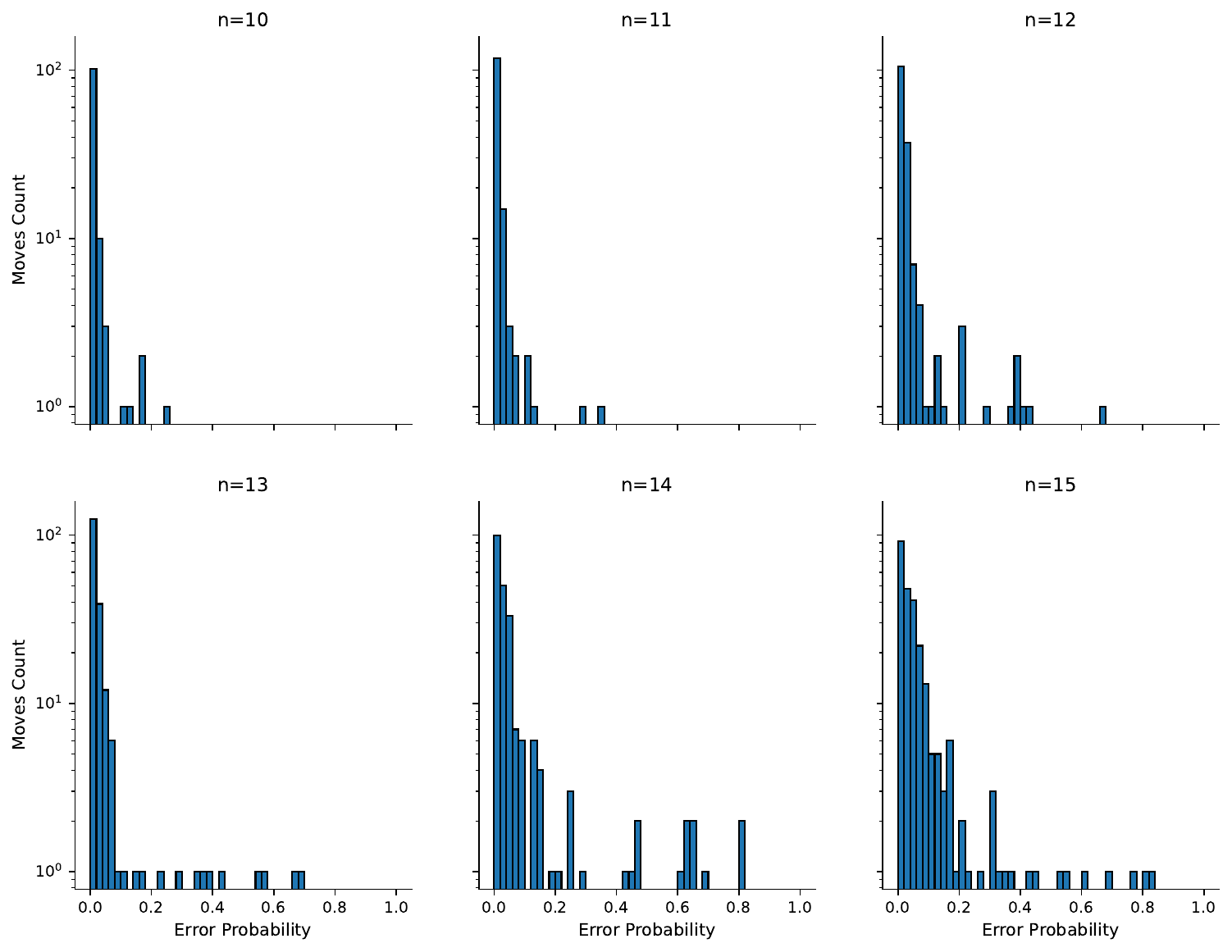}

    \caption{Error count histogram of o4-mini model for Checkers Jumping. The error counts were estimated by sampling 50 independent solutions for each step.}
\end{figure}

\begin{figure}[hb]
    \centering

    \includegraphics[width=\textwidth]{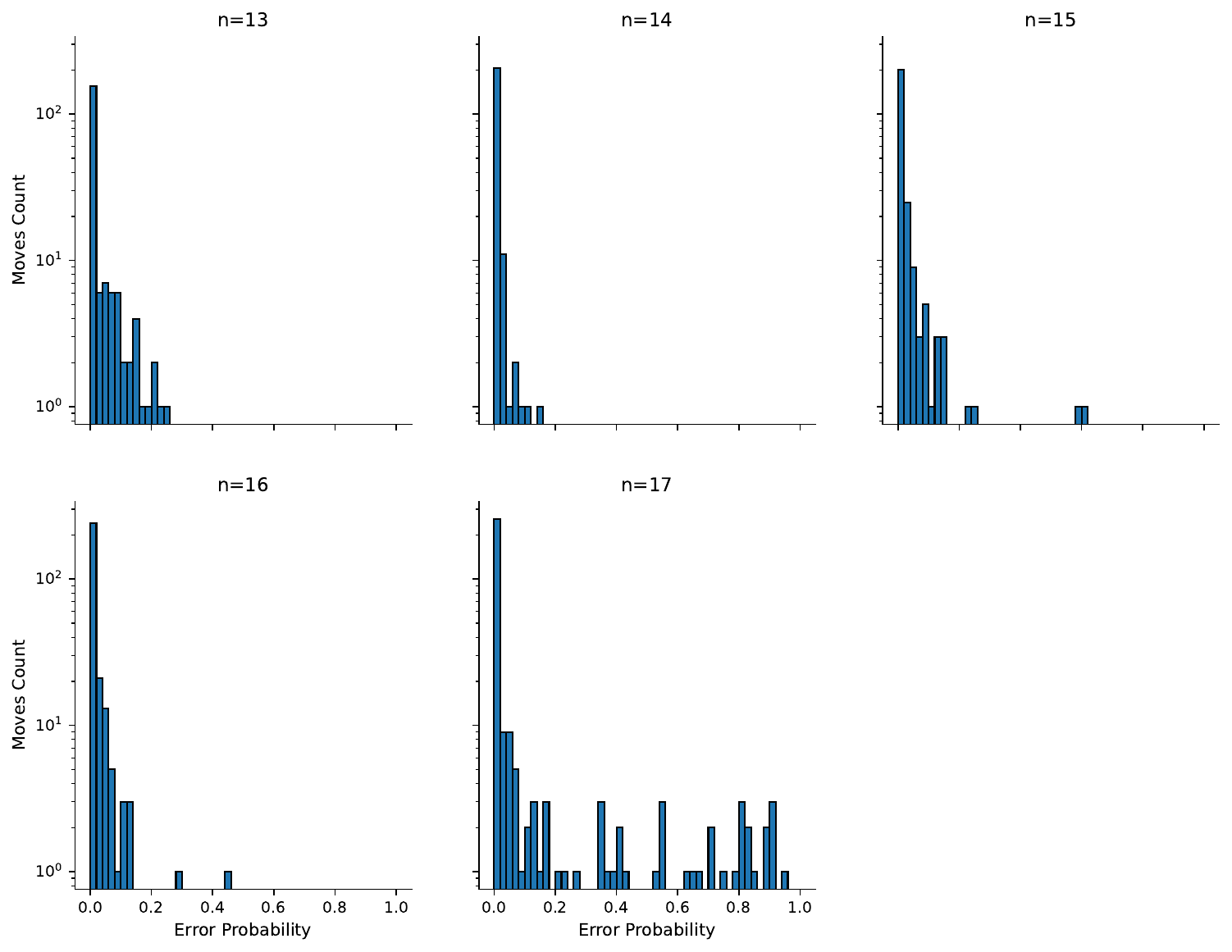}

    \caption{Error count histogram of GPT-5.2 model for Checkers Jumping. The error counts were estimated by sampling 50 independent solutions for each step.}
    \label{fig:error-hist-gpt52}
\end{figure}

\begin{figure}[hb]
    \centering

    \includegraphics[width=\textwidth]{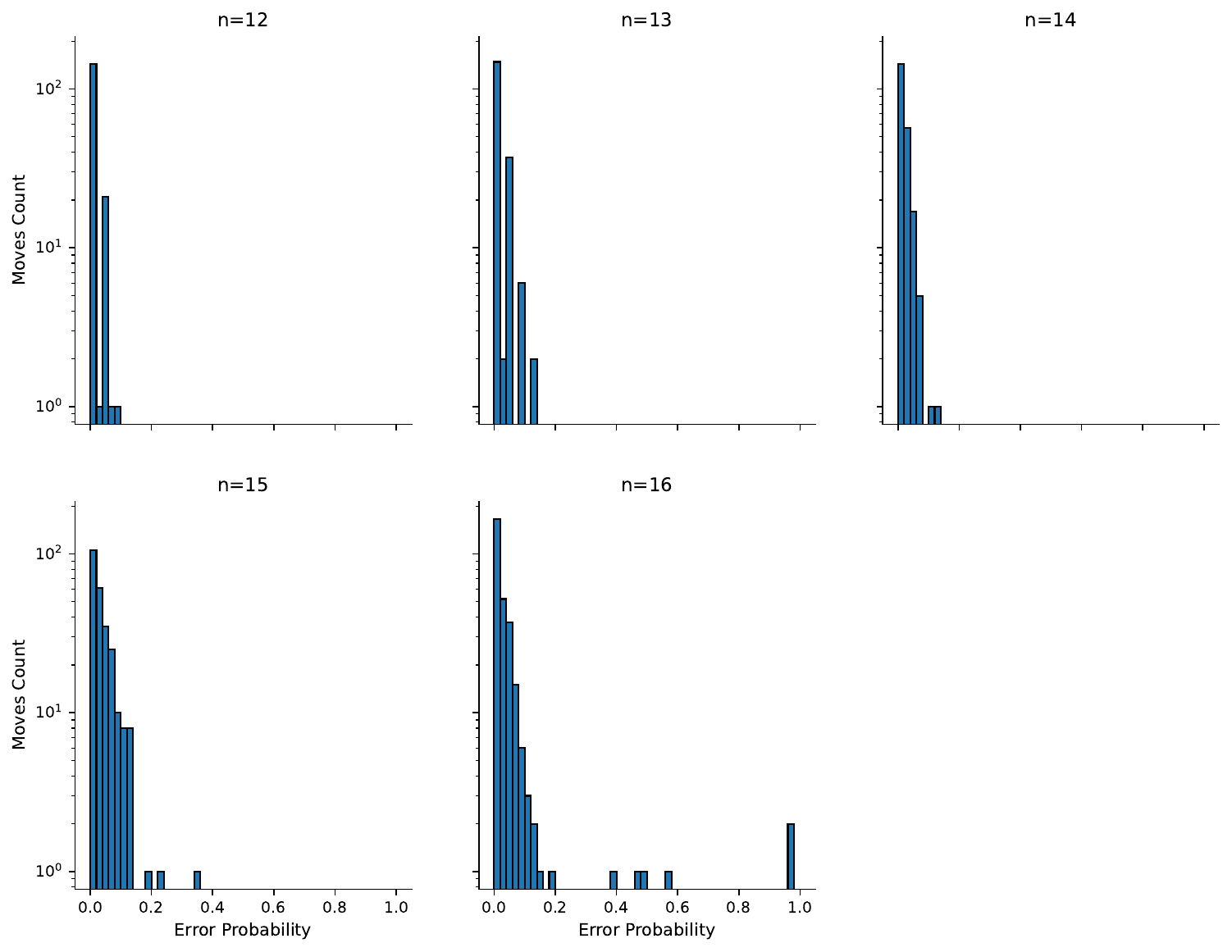}

    \caption{Error count histogram of Qwen model for Checkers Jumping. The error counts were estimated by sampling 50 independent solutions for each step.}
    \label{fig:error-hist-qwen}
\end{figure}

\begin{figure}[h]
  \centering

  \begin{subfigure}{0.48\columnwidth}
    \centering
    \includegraphics[width=\linewidth]{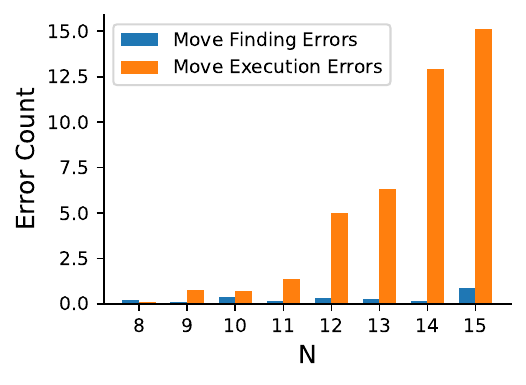}
    \caption{Checkers Jumping}
  \end{subfigure}
  \hfill
  \begin{subfigure}{0.48\columnwidth}
    \centering
    \includegraphics[width=\linewidth]{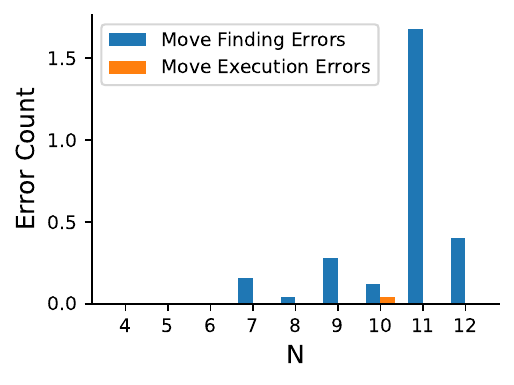}
    \caption{Tower of Hanoi}
  \end{subfigure}

  \caption{Distribution of error types for o4-mini model. Each plot shows the average number of each type errors per one solution. The main source of errors for Checkers Jumping is move execution, while for Tower of Hanoi - finding the correct move.}
  \label{fig:error-types}
\end{figure}

\clearpage\textbf{}





\section{Atomic competence barrier}

When Atomic Decomposition is combined with majority voting at each step, its failure mode converges to the dominant source of error.
As shown in Figure~\ref{fig:error-types}, execution errors—rather than incorrect move selection—are the primary failure mode on Checkers Jumping.
This suggests that the performance of Atomic Decomposition with voting is fundamentally bottlenecked by the model’s ability to reliably execute a given move.

\begin{figure}[h]
  \begin{center}
    \centering
    \includegraphics[width=\columnwidth]{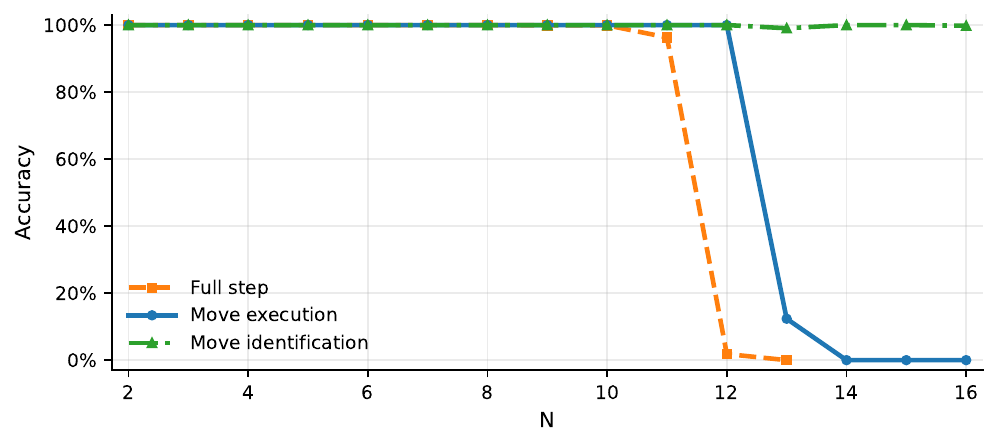}
    \caption{
    Atomic Decomposition with voting approaches the atomic competence barrier on Checkers Jumping, where the limiting factor is atomic move execution.
    The plot compares accuracy of full-step execution (selecting and executing a move) and isolated move selection or move execution as a function of problem size $n$.
    Results are shown for the o4-mini model using majority voting over 32 independently sampled solutions at each step.
    }
    \label{fig:barrier}
  \end{center}
\end{figure}

To substantiate this claim, we evaluate Atomic Decomposition with stepwise voting over $v=32$ votes under three settings:
(i) full-step execution (move finding followed by move execution),
(ii) move finding only, and
(iii) move execution only.
Figure~\ref{fig:barrier} shows that full-step execution closely approaches the performance of move execution alone, confirming that move execution constitutes the limiting factor.
We refer to this limitation as the \emph{atomic competence barrier}.

\clearpage

\section{Example of errors} \label{app-b}


\begin{lstlisting}[
  basicstyle=\ttfamily\small,
  breaklines=true,
  caption={The outputs of o4-mini model for two of the most difficult steps for this model for Checkers Jumping ($n=13$), sampled 50 times per move. The main cause of error is forgetting one red checker 'R' in a long block of consecutive red checkers, when one more red checker should join the block as a result of the move. Thus, the model outputs the board of shorter length than it should be.},
]
Step id = 168

Correct solution:
move = ['R', 17, 19]
state = ['B', 'B', 'B', 'B', 'B', 'B', 'B', 'R', 'B', 'R', 'B', 'R', 'B', 'R', 'B', 'R', 'B', '_', 'B', 'R', 'R', 'R', 'R', 'R', 'R', 'R', 'R']

Model solutions:
move = ['R', 17, 19]
state = ['B', 'B', 'B', 'B', 'B', 'B', 'B', 'R', 'B', 'R', 'B', 'R', 'B', 'R', 'B', 'R', 'B', '_', 'B', 'R', 'R', 'R', 'R', 'R', 'R', 'R']
Count: 34

move = ['R', 17, 19]
state = ['B', 'B', 'B', 'B', 'B', 'B', 'B', 'R', 'B', 'R', 'B', 'R', 'B', 'R', 'B', 'R', 'B', '_', 'B', 'R', 'R', 'R', 'R', 'R', 'R', 'R', 'R']
Count: 16

Step id = 150

Correct solution:
move = ['R', 21, 22],
state = ['B', 'B', 'B', 'B', 'B', 'R', 'B', 'R', 'B', 'R', 'B', 'R', 'B', 'R', 'B', 'R', 'B', 'R', 'B', 'R', 'B', '_', 'R', 'R', 'R', 'R', 'R']
Model solutions:

move = ['R', 21, 22],
state = ['B', 'B', 'B', 'B', 'B', 'R', 'B', 'R', 'B', 'R', 'B', 'R', 'B', 'R', 'B', 'R', 'B', 'R', 'B', 'R', 'B', '_', 'R', 'R', 'R', 'R']
Count: 33

move = ['R', 21, 22],
state = ['B', 'B', 'B', 'B', 'B', 'R', 'B', 'R', 'B', 'R', 'B', 'R', 'B', 'R', 'B', 'R', 'B', 'R', 'B', 'R', 'B', '_', 'R', 'R', 'R', 'R', 'R']
Count: 17
\end{lstlisting}

\clearpage

\begin{lstlisting}[
  basicstyle=\ttfamily\small,
  breaklines=true,
  caption={The outputs of GPT-5.2 model for two the most difficult steps for this model for Checkers Jumping ($n=13$), sampled 100 times per move. The main cause of error is adding one extra blue checker 'B' in a long block of consequtive blue checkers. Thus, the model outputs the board of bigger length than it should be.},
]
Step id = 16

Correct solution:
move = ['B', 12, 10],
state = ['R', 'R', 'R', 'R', 'R', 'R', 'R', 'R', 'B', 'R', 'B', 'R', '_', 'R', 'B', 'R', 'B', 'R', 'B', 'B', 'B', 'B', 'B', 'B', 'B', 'B', 'B']
Model solutions:

move = ['B', 12, 10],
state = ['R', 'R', 'R', 'R', 'R', 'R', 'R', 'R', 'B', 'R', 'B', 'R', '_', 'R', 'B', 'R', 'B', 'R', 'B', 'B', 'B', 'B', 'B', 'B', 'B', 'B', 'B']
Count: 78

move = ['B', 12, 10],
state = ['R', 'R', 'R', 'R', 'R', 'R', 'R', 'R', 'B', 'R', 'B', 'R', '_', 'R', 'B', 'R', 'B', 'R', 'B', 'B', 'B', 'B', 'B', 'B', 'B', 'B', 'B', 'B']
Count: 22

Step id = 17

Correct solution:
move = ['B', 14, 12],
state = ['R', 'R', 'R', 'R', 'R', 'R', 'R', 'R', 'B', 'R', 'B', 'R', 'B', 'R', '_', 'R', 'B', 'R', 'B', 'B', 'B', 'B', 'B', 'B', 'B', 'B', 'B']
Model solutions:

move = ['B', 14, 12],
state = ['R', 'R', 'R', 'R', 'R', 'R', 'R', 'R', 'B', 'R', 'B', 'R', 'B', 'R', '_', 'R', 'B', 'R', 'B', 'B', 'B', 'B', 'B', 'B', 'B', 'B', 'B']
Count: 76

move = ['B', 14, 12],
state = ['R', 'R', 'R', 'R', 'R', 'R', 'R', 'R', 'B', 'R', 'B', 'R', 'B', 'R', '_', 'R', 'B', 'R', 'B', 'B', 'B', 'B', 'B', 'B', 'B', 'B', 'B', 'B']
Count: 22
\end{lstlisting}

\clearpage

\section{More experiments with lookahead} \label{app-c}

\begin{table*}[h]
  \begin{center}
    \begin{small}
      \begin{sc}
        \begin{tabular}{lrrrrrrrr}
          \toprule
          & \multicolumn{8}{c}{\textit{Lookahead Accuracy by Position}} \\
            \cmidrule(lr){2-9}
          Step Idx  &  $+1$ & $+2$ & $+3$ & $+4$ & $+5$ & $+6$ & $+7$ & $+8$  \\
          \midrule
          60 & 1.0 & 1.0 & 1.0 & \textbf{1.0} & 1.0 & 1.0 & 1.0 & 1.0 \\
          61 & 1.0 & 1.0 & \textbf{1.0} & 1.0 & 1.0 & 1.0 & 1.0 & 1.0 \\
          62 & 0.74 & \textbf{0.74} & 0.74 & 0.74 & 0.74 & 0.74 & 0.73 & 0.73 \\
          \textbf{63} & \textbf{0.5} & 0.5 & 0.5 & 0.5 & 0.5 & 0.5 & 0.5 & 0.5 \\
          64 & 0.94 & 0.94 & 0.94 & 0.94 & 0.94 & 0.94 & 0.94 & 0.94 \\
          \bottomrule
        \end{tabular}
      \end{sc}
    \end{small}
  \end{center}
  \caption{
    Lookahead prediction accuracy for GPT-5.2 on Checkers Jumping ($n=15$),
    shown for the hardest step (step index 63) and its neighboring steps.
    Columns correspond to the position within the Lookahead rollout.
    Bold entries indicate predictions for the hardest step obtained from
    different Lookahead starting points.
    Although the hardest step is not consistently correct at its native
    position, it is predicted correctly when inferred from earlier
    Lookahead rollouts, motivating the inclusion of recent Lookahead
    predictions in the voting procedure used by LEAD.
    }
  \label{tab:lookahead-voting-ex}
\end{table*}

\begin{figure}[h]
  \centering

  \begin{subfigure}{0.48\columnwidth}
    \centering
    \includegraphics[width=\linewidth]{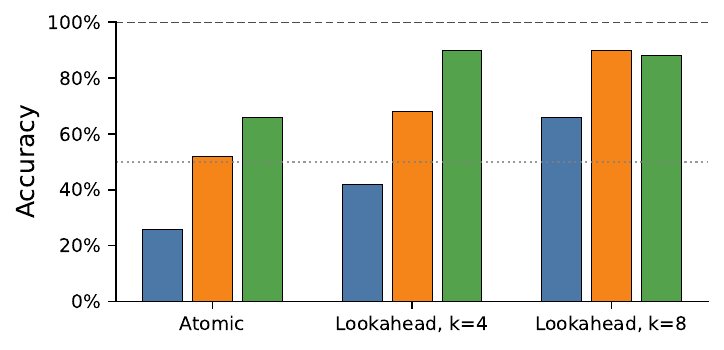}
    \caption{o4-mini}
  \end{subfigure}
  \hfill
  \begin{subfigure}{0.48\columnwidth}
    \centering
    \includegraphics[width=\linewidth]{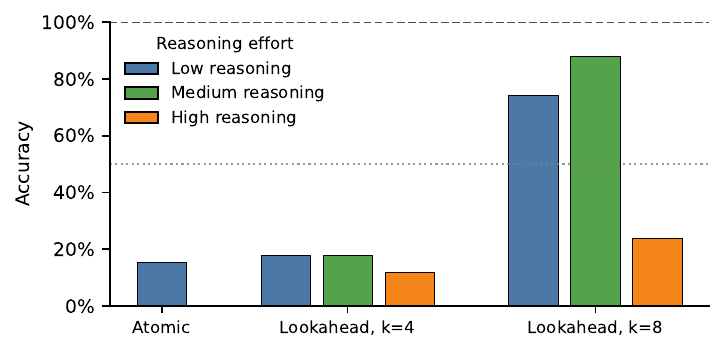}
    \caption{GPT-5.2}
  \end{subfigure}
  \caption{Comparing Lookahead with Atomic Decomposition strategies for two models on the most difficult step for that model for Checkers Jumping puzzle (the puzzle complexity is $n=13$ for o4-mini and $n=15$ for GPT-5.2 model).}
  \label{fig:lookahead-hard-steps}
\end{figure}

\clearpage

\section{Inference cost analysis} \label{app:complexity}

\begin{table}[h]
\centering
\small
\begin{tabular}{lccc}
\toprule
 & \multicolumn{3}{c}{\textit{Complexity} $n$} \\
\cmidrule(lr){2-4}
Method & 11 & 12 & 13 \\
\midrule
AD & 3.15 {\scriptsize $\pm$ 0.04} & 3.84 {\scriptsize $\pm$ 0.05} & 4.60 {\scriptsize $\pm$ 0.05} \\
AD + voting & 9.70 {\scriptsize $\pm$ 0.19} & 12.70 {\scriptsize $\pm$ 0.34} & 14.98 {\scriptsize $\pm$ 0.31} \\
LEAD & 37.73 {\scriptsize $\pm$ 0.99} & 55.96 {\scriptsize $\pm$ 1.44} & 69.84 {\scriptsize $\pm$ 1.42} \\
\bottomrule
\end{tabular}
\caption{Estimated tokens per solution ($\times 10^5$) for o4-mini}
\label{tab:complexity-o4-mini}
\end{table}

\begin{table}[h]
\centering
\small
\begin{tabular}{lccccc}
\toprule
 & \multicolumn{5}{c}{\textit{Complexity} $n$} \\
\cmidrule(lr){2-6}
Method & 13 & 14 & 15 & 16 & 17 \\
\midrule
AD & 3.18 {\scriptsize $\pm$ 0.04} & 3.64 {\scriptsize $\pm$ 0.03} & 4.40 {\scriptsize $\pm$ 0.04} & 5.16 {\scriptsize $\pm$ 0.04} & 5.91 {\scriptsize $\pm$ 0.07} \\
AD + voting & 10.11 {\scriptsize $\pm$ 0.20} & 11.07 {\scriptsize $\pm$ 0.07} & 13.95 {\scriptsize $\pm$ 0.29} & 15.86 {\scriptsize $\pm$ 0.18} & 20.73 {\scriptsize $\pm$ 0.51} \\
LEAD & 51.99 {\scriptsize $\pm$ 1.55} & 53.63 {\scriptsize $\pm$ 0.48} & 78.42 {\scriptsize $\pm$ 2.80} & 88.83 {\scriptsize $\pm$ 3.10} & 435.04 {\scriptsize $\pm$ 1.23} \\
\bottomrule
\end{tabular}
\caption{Estimated tokens per solution ($\times 10^5$) for GPT-5.2 model}
\label{tab:complexity-gpt-5.2}
\end{table}

\clearpage

\section{Solution format}

\begin{lstlisting}[caption={Expected output for Towers of Hanoi ($N = 3$)}, label={lst:task1}]
solution = [
  {'move': [1, 0, 2], 'state': [[3, 2], [], [1]]},
  {'move': [2, 0, 1], 'state': [[3], [2], [1]]},
  {'move': [1, 2, 1], 'state': [[3], [2, 1], []]},
  {'move': [3, 0, 2], 'state': [[], [2, 1], [3]]},
  {'move': [1, 1, 0], 'state': [[1], [2], [3]]},
  {'move': [2, 1, 2], 'state': [[1], [], [3, 2]]},
  {'move': [1, 0, 2], 'state': [[], [], [3, 2, 1]]},
]
\end{lstlisting}

\begin{lstlisting}[caption={Expected output for Checkers Jumping ($N = 2$)}, label={lst:task2}]
solution = [
  {'move': ['R', 0, 1], 'state': ['_', 'R', 'B']},
  {'move': ['B', 2, 0], 'state': ['B', 'R', '_']},
  {'move': ['R', 1, 2], 'state': ['B', '_', 'R']},
]
\end{lstlisting}


\clearpage

\section{Prompts} \label{sec:prompts}

\begin{lstlisting}[caption={Atomic Decomposition Prompt used for the Checkers Jumping task}, label={lst:checkers-prompt}]


Here is the puzzle.

I have a one-dimensional board with 2*N+1 cells, where N red checkers ('R') on the left, N blue checkers ('B') on the right, and one empty cell ('_') in between are arranged in a line.
The goal is to swap the positions of red and blue checkers, effectively mirroring the initial state.

Rules:
- Checkers can only move forward (towards the opposite side). More explicitly, red checkers can only move to the right, and blue checkers can only move to the left.
- A checker can slide forward into an adjacent cell, provided that this cell is empty.
- A checker can jump forward over one adjacent checker of opposite color, landing in the cell two spaces ahead (i.e., over the jumped checker), provided that this cell is empty. Note that the checker CAN NOT jump over the checker of its own color and CAN NOT jump over more than one checker.

The positions of the board are indexed from 0 (the leftmost cell) to 2*N (the rightmost cell).
We represent each move as a list of the form [<checker_color>, <position_from>, <position_to>].

Now, let's design a solution to this puzzle.

When we have a continuous block of the blue checkers adjacent to the left corner, these blue checkers have already arrived to their designated place and can not (and should not) be moved. The same with the continuous block of red checkers adjacent to the right corner. In both cases, we call such a block the ***complete block*** (in a sense that these checkers have already completed their movement).
Consequently, a blue checker is considered to be a part of a complete block, if all the cells to its left are occupied by the other blue checkers.
Similarly, a red checker is considered to be a part of a complete block, if all the cells to its right are occupied by the other red checkers.

Now, let's find out the losing position (i.e. the one for which it's impossible to reach the goal state) that we want to avoid:
 - (P1): if we have ['_', 'R', 'R'] segment on the board, and these two red checkers are not a part of a complete block, this position is losing.
 - (P2): similarly, if we have ['B', 'B', '_'] segment on the board, and these two blue checkers are not a part of a complete block, this position is losing.
 - (P3): besides, if we have ['B', '_', 'R'] segment on the board, where neither the 'B' nor the 'R' checker is a part of its complete block, this position is also losing.
Thus, when solving the puzzle, we have to avoid any move that leads to the positions (P1), (P2), or (P3).

Finally, it turns out that any valid move that does not lead to any of the losing positions (P1), (P2), or (P3), is an optimal move. It means that it's possible to solve the puzzle optimally after making such a move.

### Solution strategy
The following simple strategy will lead to the optimal solution of the puzzle: at each step, we have to select a move that does not lead to any of the losing positions (P1), (P2), or (P3).
If there multiple moves satisfying this condition, select any of them.
One may show that as long as we follow this strategy, we will always have at least one move satisfying this condition.

You DO NOT need to solve the puzzle.
Instead, you will be given an intermediate position of the game, and you have to find an optimal move for this position and execute it, outputting the subsequent state.

### Efficient Algorithm for Finding an Optimal Move and Executing It

Here's how to efficiently compute optimal moves for any board configuration:

**Step 1: Find the empty position**
- Scan the board once to find the position of '_'
- `empty_pos`: The index (0-based) of the empty cell on the board

**Step 2: Identify movement patterns around the empty cell**
Check the following patterns in O(1) time by examining cells adjacent to the empty position:

- `['R', '_']_pattern`: Boolean indicating if there's a red checker immediately to the left of the empty cell
  - Check: `board[empty_pos - 1] == 'R'` (if `empty_pos > 0`)
- `['_', 'B']_pattern`: Boolean indicating if there's a blue checker immediately to the right of the empty cell  
  - Check: `board[empty_pos + 1] == 'B'` (if `empty_pos < len(board) - 1`)
- `['R', 'B', '_']_pattern`: Boolean indicating if there's a red checker that can jump over a blue checker into the empty cell
  - Check: `board[empty_pos - 2] == 'R'` and `board[empty_pos - 1] == 'B'` (if `empty_pos > 1`)
- `['_', 'R', 'B']_pattern`: Boolean indicating if there's a blue checker that can jump over a red checker into the empty cell
  - Check: `board[empty_pos + 1] == 'R'` and `board[empty_pos + 2] == 'B'` (if `empty_pos < len(board) - 2`)

**Step 3: Generate valid moves**
Based on the patterns above, construct the list of all valid moves:

- If `['R', '_']_pattern` is True -> add `['R', empty_pos - 1, empty_pos]` (red checker slides right)
- If `['_', 'B']_pattern` is True -> add `['B', empty_pos + 1, empty_pos]` (blue checker slides left)  
- If `['R', 'B', '_']_pattern` is True -> add `['R', empty_pos - 2, empty_pos]` (red checker jumps over blue)
- If `['_', 'R', 'B']_pattern` is True -> add `['B', empty_pos + 2, empty_pos]` (blue checker jumps over red)

**Step 4: Filter optimal moves**
For each valid move, simulate the resulting board state and check if it creates any losing patterns:

For move `[color, from_pos, to_pos]`:

1. **Simulate the move:**
   - `next_state`: Copy of the current board with the move applied
     - Set `next_state[from_pos] = '_'` (source position becomes empty)
     - Set `next_state[to_pos] = color` (destination gets the checker)
   - `next_empty_pos`: Position of the empty cell after the move (equals `from_pos`)

2. **Find complete blocks in the resulting state:**
   - `next_left_block_end`: Rightmost position of the complete blue block from the left corner
     - Scan from left until first non-'B' cell, return position-1 (or -1 if no complete blue block exists)
   - `next_right_block_start`: Leftmost position of the complete red block from the right corner  
     - Scan from right until first non-'R' cell, return position+1 (or `len(board)` if no complete red block exists)

3. **Check for losing patterns:**

   **Pattern P1 `['_', 'R', 'R']`:**
   - `p1_found`: Boolean indicating if the pattern `['_', 'R', 'R']` exists starting at `next_empty_pos`
     - Check: `next_empty_pos <= len(board)-3` and `next_state[next_empty_pos+1] == 'R'` and `next_state[next_empty_pos+2] == 'R'`
   - `is_red_complete`: Boolean indicating if the two red checkers in the pattern are part of a complete red block
     - Check: `next_empty_pos+2 >= next_right_block_start` (if `p1_found` is True)
   - `p1_losing`: Boolean indicating if this pattern makes the position losing
     - Value: `p1_found and not is_red_complete`

   **Pattern P2 `['B', 'B', '_']`:**
   - `p2_found`: Boolean indicating if the pattern `['B', 'B', '_']` exists ending at `next_empty_pos`
     - Check: `next_empty_pos >= 2` and `next_state[next_empty_pos-2] == 'B'` and `next_state[next_empty_pos-1] == 'B'`
   - `is_blue_complete`: Boolean indicating if the two blue checkers in the pattern are part of a complete blue block
     - Check: `next_empty_pos-2 <= next_left_block_end` (if `p2_found` is True)
   - `p2_losing`: Boolean indicating if this pattern makes the position losing
     - Value: `p2_found and not is_blue_complete`

   **Pattern P3 `['B', '_', 'R']`:**
   - `p3_found`: Boolean indicating if the pattern `['B', '_', 'R']` exists centered at `next_empty_pos`
     - Check: `next_empty_pos >= 1` and `next_empty_pos <= len(board)-2` and `next_state[next_empty_pos-1] == 'B'` and `next_state[next_empty_pos+1] == 'R'`
   - `is_blue_complete`: Boolean indicating if the blue checker in the pattern is part of a complete blue block
     - Check: `next_empty_pos-1 <= next_left_block_end` (if `p3_found` is True)
   - `is_red_complete`: Boolean indicating if the red checker in the pattern is part of a complete red block
     - Check: `next_empty_pos+1 >= next_right_block_start` (if `p3_found` is True)
   - `p3_losing`: Boolean indicating if this pattern makes the position losing
     - Value: `p3_found and not is_blue_complete and not is_red_complete`

4. **Determine if move is optimal:**
   - The move is optimal if `not (p1_losing or p2_losing or p3_losing)`

**Step 5: Execute the first optimal move**
Once all valid moves have been filtered:

1. **Select the move:**
   - `selected_move`: The first move from the `optimal_moves` list
     - Format: `[color, from_pos, to_pos]`

2. **Execute the move:**
   - `new_state`: Copy of the current board
   - Set `new_state[from_pos] = '_'` (source position becomes empty)
   - Set `new_state[to_pos] = color` (destination gets the checker)

3. **Return the result:**
   - Return a dictionary with the selected move and the resulting board state
   - Format: `{'move': selected_move, 'new_state': new_state}`

You will be given a position of the board, and your task is to find an optimal move for this position and execute it, outputting the subsequent state (see example below).

For instance, assume N = 2 and the following position is given:

position = ['R', '_', 'R', 'B', 'B']

The solution for this query is the following:

solution = {'move': ['B', 3, 1], 'new_state': ['R', 'B', 'R', '_', 'B']}

### Task
Consider this puzzle for N = 3 and the following position of the board:

position = ['B', 'B', 'B', 'R', '_', 'R', 'R']

Find an optimal move for this position and the new state after executing this move.

Requirements
- Your final answer must be in the following format:

solution = {'move': <move>, 'new_state': <new state>}

\end{lstlisting}

\begin{lstlisting}[caption={Iterative Restart prompt ending used for the Checkers Jumping task}, label={lst:checkers-prompt-iterative}]
...

### Task
Consider this puzzle for N = 3 and the following position of the board:

position = ['R', 'R', 'R', '_', 'B', 'B', 'B']

Continue the solution from the given position.
If you can complete the solution, output the entire solution completion. Otherwise, output the partial solution that you were able to find.

Requirements
- Your final answer must be in the following format:

moves = [
  {'move': [<checker_color>, <position_from>, <position_to>], 'state': <subsequent state>},
  ...
]

- Do not include any explanatory comments in the final output.
\end{lstlisting}

\begin{lstlisting}[caption={Atomic Decomposition prompt  used for the Tower of Hanoi task}, label={lst:tower-hanoi}]
Here is the puzzle.

There are three pegs and N disks of different sizes stacked on the first peg from the largest (on the bottom) to the smallest (on top). 
The disks are numbered from 1 (smallest) to N (largest). We denote the pegs as peg_0, peg_1, and peg_2, where peg_0 is the first (starting peg), and peg_2 is the third (destination) peg.

Rules:
1. Only one disk can be moved at a time.
2. Each move consists of taking the upper disk from one stack and placing it on top of another stack.
3. A larger disk may not be placed on top of a smaller disk.

The goal is to move the entire stack to the third peg.

We represent each step as a dict with the following items:
- "step_id": int, index of the current move.
- "move": a list of the form [disk_number, peg_from, peg_to]
  - disk_number -> the number of the disk being moved (1 = smallest)
  - peg_from -> the index (0-2) of the peg the disk is moved from
  - peg_to -> the index (0-2) of the peg the disk is moved to
- "state": a list of three lists representing the configuration of the pegs after the move
  - Each inner list represents one peg (peg 0, peg 1, peg 2)
  - Each peg list contains the disks currently on that peg, ordered from bottom (on the left) to top (on the right)

You DO NOT need to solve the whole puzzle.
Instead, you will be given an intermediate position in the game (together witg the last step index), and your task is to output the next step (see example below).

For instance, assume N = 3 and the following configuration is given:

configuration = {'step_id': 3, 'state': [[3], [2, 1], []]}

The solution for this query is the following:

solution = {'step_id': 4, 'move': [3, 0, 2], 'old_state': [[3], [2, 1], []], 'state': [[], [2, 1], [3]]}

### Iterative Solution for Tower of Hanoi

Here is the algorithm how to perform the next step in the solution, given the current configuration as follows:

configuration = {step_id: <step_id>, state: <state>}

Algorithm:

1. Calculate number of completed moves:
    num_moves = step_id

2. Find a correct move:

    If num_moves is even:
        - Find the peg which contains disk 1 on its top. Denote this peg as peg_i
        - Define the movement cycle for the smallest disk based on the parity of N (total number of disks):
            - If N is odd: cycle = [peg_0, peg_2, peg_1]
            - If N is even: cycle = [peg_0, peg_1, peg_2]
        - Let peg_j be the next peg after peg_i in cycle (in cyclic order, meaning that the the next peg after cycle[-1] is cycle[0])
        - Move disk 1 from peg_i to peg_j:
            move = [1, peg_i, peg_j]

    If num_moves is odd:
        - Sort the three pegs by the size of their top disk in descending order (treating an empty peg as having an infinitely large top disk).
            Denote sorted pegs as peg_i, peg_j peg_k (meaning that peg_i has the largest disk on top, peg_k - the smallest).
            Let d be the number of the top disk on peg_j.
        - Move disk d from peg_j to peg_i:
            move = [d, peg_j, peg_i]

3. Execute the move:
    - Let the found move be 
        move = [<disk_number>, <peg_from>, <peg_to>]
    - Update state:
        - create new_state as a copy of state
        - remove the last element from new_state[peg_from] list
        - append <disk_number> to new_state[peg_to] list

4. Update the step_id:
    - new_step_id = step_id + 1

5. Output the new step:
    solution = {'new_step_id': new_step_id, 'move': move, 'new_state': new_state}


# Task:
Consider this puzzle for N = 4 and the following configuration of the game:

configuration = {step_id: 0, state: [[4, 3, 2, 1], [], []]}

Requirements:
- Your final answer must be in the following format:

solution = {'new_step_id': <new_step_id>, 'move': <move>, 'new_state': <new_state>}

\end{lstlisting}
\end{document}